\documentclass[letterpaper, 10 pt, conference]{ieeeconf}  
\pdfoutput=1
\IEEEoverridecommandlockouts                              

\usepackage[T1]{fontenc}
\usepackage[utf8]{inputenc} 
\usepackage{amsmath} 
\usepackage{amssymb}  
\usepackage{dsfont}
\usepackage{bm}
\usepackage{wrapfig}

\usepackage{graphicx} 

\usepackage{algorithm}
\usepackage[noend]{algpseudocode}
\usepackage{subfigure}
\usepackage{color}
\usepackage{url}

\floatname{algorithm}{Procedure}
\renewcommand{\algorithmicrequire}{\textbf{Input:}}
\renewcommand{\algorithmicensure}{\textbf{Output:}}

\DeclareMathOperator*{\argmax}{\arg\!\max}

\makeatletter
\newcommand\footnoteref[1]{\protected@xdef\@thefnmark{\ref{#1}}\@footnotemark}
\makeatother

\title{\LARGE \bf
Bayesian Optimization with Automatic Prior Selection\\for Data-Efficient Direct Policy Search
}

\author{R\'emi Pautrat, Konstantinos Chatzilygeroudis and Jean-Baptiste Mouret$^{*}$
\thanks{*Corresponding author: {\tt\footnotesize jean-baptiste.mouret@inria.fr}}
\thanks{\scriptsize All authors have the following affiliations:}
\thanks{\scriptsize - Inria, Villers-lès-Nancy, F-54600, France}%
\thanks{\scriptsize - CNRS, Loria, UMR 7503, Vandœuvre-lès-Nancy, F-54500, France}%
\thanks{\scriptsize - Université de Lorraine, Loria, UMR 7503, Vandœuvre-lès-Nancy, F-54500, France}%
\thanks{\scriptsize This work received funding from the European Research Council (ERC) under the European Union's Horizon 2020 research and innovation programme (GA no. 637972, project ``ResiBots'') and the European Commission through the project H2020 AnDy (GA no. 731540).}%
}

\begin{document}

\maketitle
\thispagestyle{empty}
\pagestyle{empty}


\begin{abstract}
	 One of the most interesting features of Bayesian optimization for direct policy search is that it can leverage priors (\emph{e.g.}, from simulation or from previous tasks) to accelerate learning on a robot. In this paper, we are interested in situations for which several priors exist but we do not know in advance which one fits best the current situation. We tackle this problem by introducing a novel acquisition function, called Most Likely Expected Improvement (MLEI), that combines the likelihood of the priors and the expected improvement. We evaluate this new acquisition function on a transfer learning task for a 5-DOF planar arm and on a possibly damaged, 6-legged robot that has to learn to walk on flat ground and on stairs, with priors corresponding to different stairs and different kinds of damages. Our results show that MLEI effectively identifies and exploits the priors, even when there is no obvious match between the current situations and the priors.
\end{abstract}


\section{Introduction}
	Reinforcement Learning (RL)~\cite{sutton1998reinforcement} could allow robots to adapt to new tasks (\emph{e.g.}, a new tool) and new contexts (\emph{e.g.}, a damage~\cite{NatureArticle,chatzilygeroudis2016reset}), but only if this adaptation happens in a few minutes: contrary to simulated worlds (\emph{e.g.}, games), where thousands (if not millions) of simulations can be evaluated, the number of trials in robotics hardware is limited by the energetic autonomy of the robot and the need to perform the task as soon as possible to be useful~\cite{mouret2016micro}.

	Among the different approaches to data-efficient RL, Bayesian Optimization (BO) is a promising approach because it can work with continuous action and state spaces, contrary to classic RL algorithms~\cite{Deisenroth}, and because it scales well with the dimension of the state space, contrary to model-based policy search algorithms (\emph{e.g.}, PILCO~\cite{PILCO} or Black-DROPS~\cite{chatzilygeroudis2017black}). For example, BO was successfully used to learn walking policies for a quadruped~\cite{BOgait} and for a 2-legged compass walker~\cite{calandra2016bayesian}.

	BO was originally conceived as a black-box optimization algorithm for expensive functions~\cite{jones1998efficient,shahriari2016taking}. However, in robot learning, it is often possible to have some prior knowledge about the behavior of the system. For instance, a simulator of an intact robot can help to learn a policy on a damaged robot~\cite{NatureArticle} or to guide the search algorithm to the most promising areas~\cite{cutler2015efficient,ko2007gaussian}; or knowledge acquired when solving previous tasks can make it faster to solve a new task (transfer learning)~\cite{taylor2009transfer}. When BO is used for direct policy search, priors on the reward function can be added by using a non-constant mean function in the model, that is, by modeling the difference between the observations and the prior instead of modeling the observations directly~\cite{ko2007gaussian,NatureArticle}.

	In this paper, we are interested in using BO when (1) several priors are available and, (2) we do not know beforehand which prior corresponds to the current context. A typical situation is a robot that knows how to solve a task in context A, B, and C (priors) and needs to learn to solve it in context D, while not knowing whether D is closer to A, B, or C. For some tasks, a perception system might recognize the right context~\cite{plagemann}, but in many others only the observations of the reward function can allow the robot to determine what prior is the most plausible. For instance, a walking robot could learn that a surface is slippery by observing that it matches the predictions that correspond to a prior for slippery floors, but it is often difficult to predict the slipperiness of a surface by only looking at it.

	Our main insight is that we can compare two priors by computing the likelihood of the combination "prior + model" so that we can select the prior that matches the best the observations. Our second insight is that this prior selection can be elegantly incorporated as an acquisition function of a BO procedure, so that we select the next point to test by balancing between the expected improvement and the likelihood of the model used to compute the expected improvement. We demonstrate our approach on a simple simulated arm problem whose goal is to reach a target and on a simulated and physical 6-legged robot that faces different damage conditions and different environments.


\section{Related work}

\subsection{Direct policy search in robotics}

Direct policy search is a successful approach for RL in robotics because it scales well to high dimensional and continuous state-action spaces~\cite{Deisenroth,arulkumaran2017brief}.
Instead of trying to predict the expected returns of future events with value-function based learning as in TD learning~\cite{Sutton1988}, direct policy search algorithms look for for the optimal parameters of parameterized policies. They essentially differ in the way the policy is updated, with techniques ranging from gradient estimation with finite differences~\cite{kohl2004policy} to more advanced optimization algorithms such as the Covariance Matrix Adaptation Evolutionary Strategy (CMA-ES)~\cite{Hansen01acompletely,stulp2013robot}, Trust Region Policy Optimization (TRPO)~\cite{schulman2015trust} or Deep Deterministic Policy Gradient (DDPG)~\cite{lillicrap_continuous_2015} algorithms. To make learning tractable, most of the successful experiments rely on prior knowledge through demonstrations~\cite{Deisenroth} and on low-dimensional policy representations (\emph{e.g.}, dynamic movement primitives~\cite{stulp2013robot}): without such hand-designed priors, thousands of episodes are usually required~\cite{stulp2013robot,Deisenroth,polydoros2017survey}.

Model-based policy search is an alternative to direct policy search that aims at improving the data-efficiency, that is, to minimize the number of required trials~\cite{PILCO,chatzilygeroudis2017black}. To do so, model-based policy search algorithms choose the next policy by: (1) performing an episode on the robot, (2) learning a dynamical model of the system using the data acquired so far, and (3) optimizing the policy \emph{according to the model} using a direct policy search algorithm. These algorithms scale well with the dimensionality of the policy (the number of parameters to optimize) because the policy optimization is performed on the model; but they are very sensitive to the dimension of the state-space because they need to learn to predict accurately the next state given the current one. Their difficulty to scale up makes it challenging to use them for systems that are more complex than basic control benchmarks (\emph{e.g.}, cart-pole or simple manipulators)~\cite{Deisenroth,polydoros2017survey}.

\subsection{Bayesian optimization for RL}
Instead of modeling the dynamics of the system, Bayesian Optimization (BO) directly models the reward function~\cite{Brochu, shahriari2016taking}; it then leverages this model to predict the most promising set of parameters for the policy, that is, those that maximize the expected reward. After each episode, BO updates the model, which allows it to improve the predictions for the next iteration.

The core of Bayesian optimization is made of two main components: a model of the reward function, and an \emph{acquisition function}, which uses the model to define the utility of each point of the search space. The vast majority of experiments with Bayesian optimization use Gaussian Processes (GP)~\cite{Rasmussen} as a model. For the acquisition function, most of them use the Expected Improvement (EI), the Upper Confidence Bound (UCB) or the Probability of Improvement (PI)~\cite{Brochu,hennig2011}. Experimental results tend to show that EI can perform better on artificial objective functions than PI and UCB~\cite{hennig2011}, but a recent experiment on gait learning on a physical robot suggested that UCB can outperform EI in real situations~\cite{calandra2016bayesian}.

As a direct policy search approach, BO does not depend on the dimensionality of the state space, which makes it effective for learning policies for robots with complex dynamics (\emph{e.g.}, locomotion, because of the non-linearity created by the contacts). For instance, Bayesian optimization was successfully used to learn policies for a quadruped robot~\cite{BOgait}, a quadcopter~\cite{berkenkamp16safe}, a small biped ``compass robot''~\cite{calandra2016bayesian}, or a pocket-sized soft tensegrity robot~\cite{rieffel2017soft}. In all of these cases, BO was at least an order of magnitude more data-efficient than competing methods. In a different domain, BO is also becoming one of the most successful approaches to tune the hyper-parameters of machine learning algorithms~\cite{shahriari2016taking}: like in robotics, evaluating the quality of each set of parameters takes a long time.

\subsection{Priors for Gaussian processes}
One benefit of using GP as a modeling method is that we can easily include prior knowledge about the data. The most common way is to select a particular \emph{mean function}, which roughly corresponds to ``what is the predicted value when there is no data?''.

Early work on Bayesian optimization for robotics focused on constant mean functions \cite{BOgait} (\emph{i.e.}, $\mu(x) = C$ where $C$ is a user-defined constant). They noted that an overestimating mean function will make the real data appear mediocre, which leads to an excessive exploration, whereas an underestimating prior will lead to a greedy exploration since all the real observations will look promising~\cite{BOgait}.

More recent work proposed priors that come from simulators or simplified models, that is, non-constant priors. In particular, the ``Intelligent Trial \& Error'' (IT\&E) algorithm~\cite{NatureArticle,papaspyros2016safety} first creates a repertoire of about $15,000$ high-performing policies and stores them in a low-dimensional map (\emph{e.g.}, 6-dimensional whereas the policy space is 36-dimensional). When the robot needs to adapt, a Bayesian optimization algorithm searches for the best policy in the low-dimensional map and uses the reward stored in the map as the mean function of a GP. This algorithm allowed a 6-legged walking robot to adapt to several damage conditions (\emph{e.g.}, a missing leg or a shortened leg) in less than 2 minutes (less than a dozen of trials), whereas it used a simulator of the intact robot to generate the prior. Instead of generating the prior first, it is also possible to choose between querying the simulator or the real robot~\cite{marco17virtualvsreal} and add the point with a different ``confidence level'' (noise) depending on how they were obtained. Last, a recent article proposed to use a simulator to learn the kernel used in the GP, instead of using a simulator to define the mean function \cite{antonova2017deep}.

Priors from simulation were also successfully used in model learning with GPs: instead of learning the dynamical model of the robot from scratch, it is possible to learn a ``residual model'', that is, the difference between the simulated and the real robot~\cite{ko2007gaussian,lee2017gp,saverianodata}. This approach was, for instance, successfully demonstrated with the PILCO algorithm for model-based policy search~\cite{cutler2015efficient,saverianodata} and when learning a model for optimal control~\cite{lee2017gp}.

While these contributions show that using well-chosen priors with GPs is a promising approach for data-efficient learning, all the previous algorithms assume that we know the ``right'' prior in advance. This is often a strong assumption because it means that the robot recognizes the current situation; this is also often a critical assumption because a misleading prior can substantially slow down the learning process. In the present paper, we relax this assumption by allowing the algorithm to choose the prior that is the most likely to help the learning process. For instance, we can have priors that correspond to different typical situations and let the algorithm choose automatically the most relevant one (and ignore the misleading ones).


\section{Combining Likelihood and Expected Improvement}

Like in most BO implementations, we model the objective function $F(\bm{x})$ to be maximized over the space $\mathcal{X}$ by a Gaussian process $f(\bm{x})$ with a mean function $m(\cdot)$ and a covariance function $\kappa(\cdot,\cdot)$:
	$$f(\bm{x}) \sim \mathcal{GP}(m(\bm{x}), \kappa(\bm{x}, \bm{x'}))$$

Let us assume that we already made $t$ observations on the points $\bm{x}_1, ..., \bm{x}_t \in \mathcal{X}$ (abridged as $\bm{x}_{1..t}$) that are summed up in the vector $\bm{F}(\bm{x}_{1..t}) = (F(\bm{x}_1), ..., F(\bm{x}_t))$, and that we fixed a noise parameter $\sigma_{n}$. The GP for a new point $\bm{x}\in\mathcal{X}$ is computed using a kernel function $\kappa(\bm{x}, \bm{y})$, a kernel vector $\bm{k}$, and a kernel matrix $\bm{K}$~\cite{Rasmussen}:
	\begin{align}
	&P(f(\bm{x}) ~|~ \bm{x}_{1..t}) \sim \mathcal{N}(\mu_t(\bm{x}), \sigma_t^2(\bm{x})), \mathrm{where: } \\
	&\mu_t(\bm{x}) = \bm{k}^T \bm{K}^{-1} \bm{F}(\bm{x}_{1..t}) \\
	&\sigma_t^2(\bm{x}) = \kappa(\bm{x}, \bm{x}) - \bm{k}^T \bm{K}^{-1} \bm{k} \\
	&\bm{K} = \begin{bmatrix}
		\kappa(\bm{x}_1, \bm{x}_1) & \dots & \kappa(\bm{x}_1, \bm{x}_t) \\
		\vdots & \ddots & \vdots \\
		\kappa(\bm{x}_t, \bm{x}_1) & \dots & \kappa(\bm{x}_t, \bm{x}_t)
	\end{bmatrix} + \sigma_{n}^2 \bm{I} \\
	&\bm{k} = \begin{bmatrix}
		\kappa(\bm{x}, \bm{x}_1) & \dots & \kappa(\bm{x}, \bm{x}_t)
	\end{bmatrix}^T
	\end{align}

In many situations, some prior knowledge on the objective function is available before starting the optimization. In that case, we can write this information with a prior function $\mathcal{P}$ and update the equations of the GP accordingly~\cite{NatureArticle}:
	\begin{align}
	\mu_t(\bm{x}) &= \mathcal{P}(\bm{x}) + \bm{k}^T \bm{K}^{-1} (\bm{F}(\bm{x}_{1..t}) - \bm{\mathcal{P}}(\bm{x}_{1..t}))
	\end{align}

The next point $\bm{x}$ where the objective function should be evaluated is found by maximizing an \emph{acquisition function}, that is, a function that leverages the model (both the variance and the mean) to predict the most promising point. A function that is often used for this is the Expected Improvement (EI)~\cite{OptimalLearning,shahriari2016taking}:
	\begin{align}
	\begin{split}
	&\mathrm{EI}(\bm{x}) = \mathbb{E}(\mathrm{I}(\bm{x})) = \mathbb{E}(\max(0,f(\bm{x}) - M_t))  \\
		  &= \begin{cases}
		(\mu_t(\bm{x}) - M_t) \Phi(Z) + \sigma_t(\bm{x}) \phi(Z) & \text{if } \sigma_t(\bm{x}) \neq 0 \\
                  0 & \text{if } \sigma_t(\bm{x})=0
		\end{cases}
	\end{split}
    \end{align}
    where $M_t=\max_{i=1..t} F(\bm{x_i})$ is the best value observed at time t, $Z = \frac{\mu_t(\bm{x}) - M_t}{\sigma_t(\bm{x})}$, $\Phi$ and $\phi$ are respectively the cumulative and probability density functions of the standard normal distribution.

Choosing the best prior can be seen as a problem of model selection (since the prior is part of the model), which is effectively achieved by comparing the likelihood of alternative models~\cite{Rasmussen}:
	\begin{align}
	 &P(\bm{f}(\bm{x}_{1..t}) ~|~ \bm{x}_{1..t},\bm{\mathcal{P}}(\bm{x}_{1..t}))=\nonumber\\
	 &\frac{1}{\sqrt{(2\pi)^t |K|}} \exp\Big(-\frac{1}{2} \big(\bm{F}(\bm{x}_{1..t}) \cdots \nonumber\\
	 & \cdots - \bm{\mathcal{P}}(\bm{x}_{1..t}))^T \bm{K}^{-1} (\bm{F}(\bm{x}_{1..t}) - \bm{\mathcal{P}}(\bm{x}_{1..t})\big)\Big)
\end{align}

Intuitively, we could select the prior that corresponds to the best likelihood, then compute the expected improvement for this model. However, we would risk to select an ``over-pessimistic'' prior at the beginning of the optimization, because the first observations (which are often random points) are likely to be low-performing --- if random points were likely to be high-performing, there would be no need for learning. In essence, if we have not yet observed any high-performing solutions, then the likeliest prior is a prior for which every solution is low-performing.

We therefore need to balance between the likelihood of the prior and the potential for high-performing solutions. In other words, a good expected improvement according to an unlikely model should be ignored; conversely, a likely model with a low expected improvement might be too pessimistic (``nothing works'') and not helpful. 
A model that is ``likely enough'' and lets us expect some good improvement might be the most helpful to find the maximum of $F$.

Let us assume that the objective function only takes discrete values, in which case the likelihood is a probability. Considering $t$ observations $F(\bm{x}_1), ..., F(\bm{x}_t)$, we introduce the indicator function $\mathds{1}_{f(\bm{x}_1)=F(\bm{x}_1), ..., f(\bm{x}_t)=F(\bm{x_t)}}$ which equals to $1$ when the predictions match exactly the observations, and we define the Expected Improvement for a prior $\mathcal{P}$:
	\begin{align}
	&\mathrm{EIP}(\bm{x}, \mathcal{P}) = \mathbb{E}(\mathrm{I}(\bm{x}) \times \mathds{1}_{f(\bm{x}_1)=F(\bm{x}_1), ..., f(\bm{x}_t)=F(\bm{x}_t)}) \\
						& = \mathbb{E}(\max(0,f(\bm{x}) - M_t ) \times \mathds{1}_{f(\bm{x}_1)=F(\bm{x}_1), ..., f(\bm{x}_t)=F(\bm{x}_t)} ) \nonumber
	\end{align}

But as the predicted value $f(\bm{x}) \sim \mathcal{N}(\mu_t(\bm{x}), \sigma_t^2(\bm{x}))$ only depends on the samples $\bm{x}_1, ..., \bm{x}_t$, the observations $\bm{F}(\bm{x}_{1..t})$ and the deterministic function $\mathcal{P}$,  it is independent of the original distribution $\bm{f}(\bm{x}_{1..t}) \sim \mathcal{N}(\bm{\mathcal{P}}(\bm{x}_{1..t}), \bm{K})$. Thus the two factors inside the expectation are two independent variables and can be split:
	\begin{align}
	\mathrm{EIP}(\bm{x}, \mathcal{P}) & = \mathbb{E}( \max(0,f(\bm{x}) - M_t)) \nonumber \\
						& ~ ~ ~ \times \mathbb{E}( \mathds{1}_{f(\bm{x}_1)=F(\bm{x}_1), ..., f(\bm{x}_t)=F(\bm{x}_t)} ) \\
						& = \mathrm{EI}(\bm{x}) \times P(~\bm{f}(\bm{x}_{1..t}) \textit{\small{=}} \bm{F}(\bm{x}_{1..t}) ~|~ \bm{x}_{1..t}, \mathcal{P}) \nonumber
	\end{align}

	This new function can be extended afterwards to the case where $F$ takes continuous values: the likelihood becomes a density probability function, but the $\mathrm{EIP}$ can still be defined as the product of the expected improvement with the likelihood:
	\begin{equation}
	\mathrm{EIP}(\bm{x}, \mathcal{P}) = \mathrm{EI}(\bm{x}) \times P(\bm{f}(\bm{x}_{1..t}) ~|~ \bm{x}_{1..t}, \bm{\mathcal{P}}(\bm{x}_{1..t}))
	\end{equation}

	When we have $m$ priors $\mathcal{P}_1, \cdots, \mathcal{P}_m$, the Most Likely Expected Improvement (MLEI) acquisition function can then be defined as:
	\begin{equation}
	\mathrm{MLEI}(\bm{x}, \mathcal{P}_1, \cdots, \mathcal{P}_m) = \max_{p \in \mathcal{P}_1, \cdots, \mathcal{P}_m} \mathrm{EIP}(\bm{x}, p)
	\end{equation}

The MLEI acquisition function can be used like any other acquisition function in the BO algorithm. Please note that the likelihood has to be evaluated only once for each model (that is, once for each prior), and not for every point $\bm{x}$ (see Algo.~\ref{algo}). We use the C++-11 Limbo library for the BO implementation~\cite{cully2016limbo}.

	\begin{algorithm}[H]
		\caption{Bayesian Optimization with MLEI}
		\label{BOmPriors}
		\label{algo}
		\begin{algorithmic}[1]
		\Procedure{BOMultiplePriors}{} \\
		\algorithmicrequire ~$m$ priors $\mathcal{P}_1, ..., \mathcal{P}_m$, an objective function $F$ \\
		\algorithmicensure ~An approximation of the maximum of $F$
		\State Initialize $m$ Gaussian processes $f_1, ..., f_m$ with the $m$ priors and the kernel function $\kappa$:
		\State $\forall i \in \{1, ..., m\}, f_i(\bm{x}) \sim \mathcal{N}(\mathcal{P}_i(\bm{x}), \kappa(\bm{x}, \bm{x'}))$
		\State $t \gets 1$
		\While{$t\leq maxIterations$}
			\For{$i=1..m$}
				\State $l \gets \textit{computeLikelihood}(f_i, \bm{x}_1, ..., \bm{x}_{t-1})$
				\State $\bm{s}_i \gets \argmax_{\bm{x}\in\mathcal{X}} \mathrm{EI}(\bm{x})$
				\Statex\hspace{6em}$= \argmax_{\bm{x}\in\mathcal{X}} (\mathbb{E}(\max(0,f_i(\bm{x}) - M_t)))$
				\State $\mathrm{EIP}(\bm{s}_i, \mathcal{P}_i) \gets l \times \mathrm{EI}(\bm{s}_i)$
			\EndFor
			\State $\bm{x}_t, p_t \gets \argmax_{i=1..m} \mathrm{EIP}(\bm{s}_i, \mathcal{P}_i)$
			\State Evaluate $F(\bm{x}_t)$ on the robot
			\State Update the $m$ Gaussian processes with the new observation $F(\bm{x}_t)$
			\State $t \gets t+1$
		\EndWhile
		\Return $\max_{t=1..maxIterations} F(\bm{x}_t)$
		\EndProcedure
		\end{algorithmic}
	\end{algorithm}


\section{Experimental Results}

\subsection{Robotic arm experiment (transfer learning)}

\begin{figure}
	\centering
	\subfigure[]{\label{arm_setup} \includegraphics[width=0.22\textwidth]{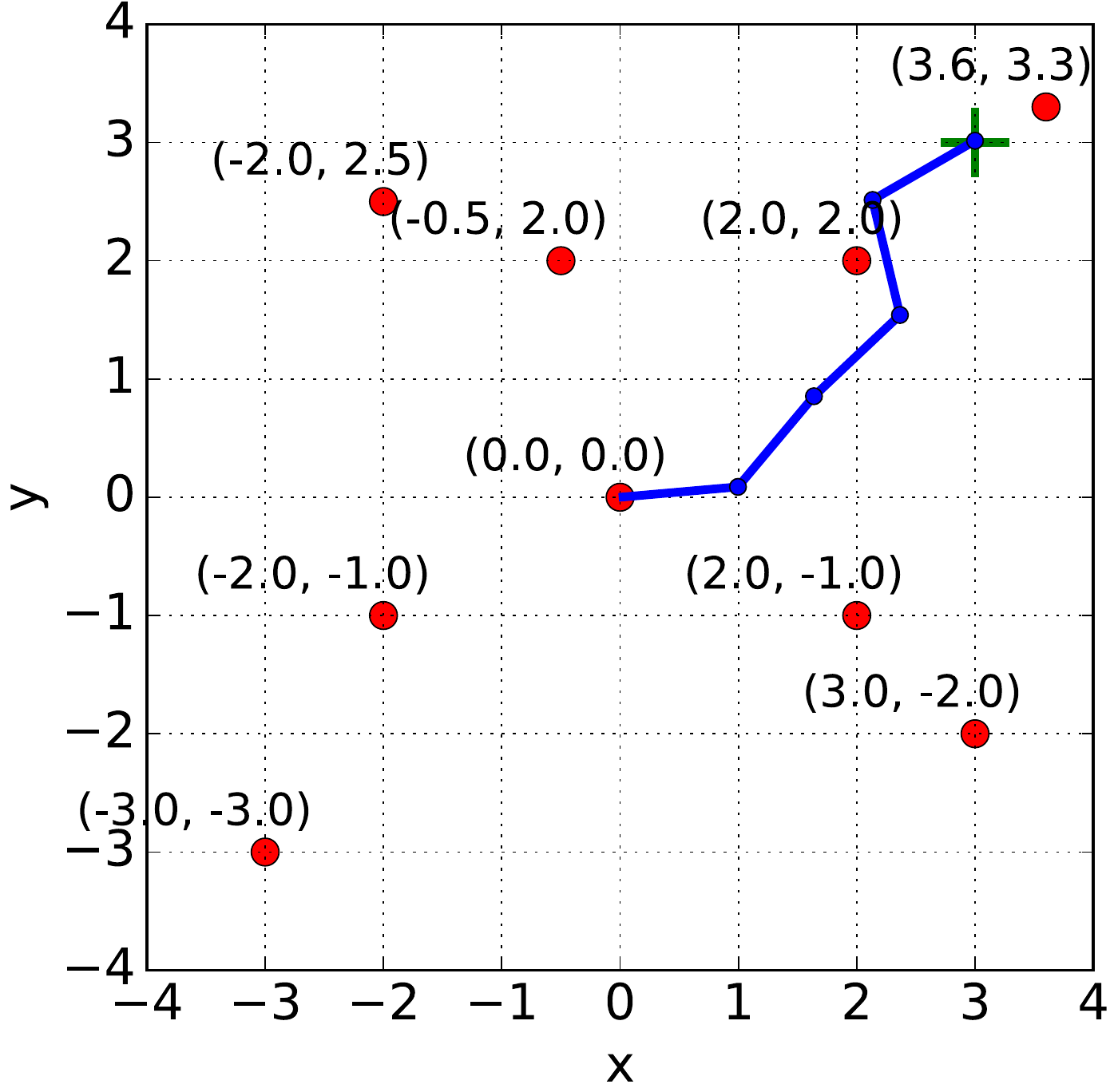}}
	\subfigure[]{\label{arm_results} \includegraphics[width=0.24\textwidth]{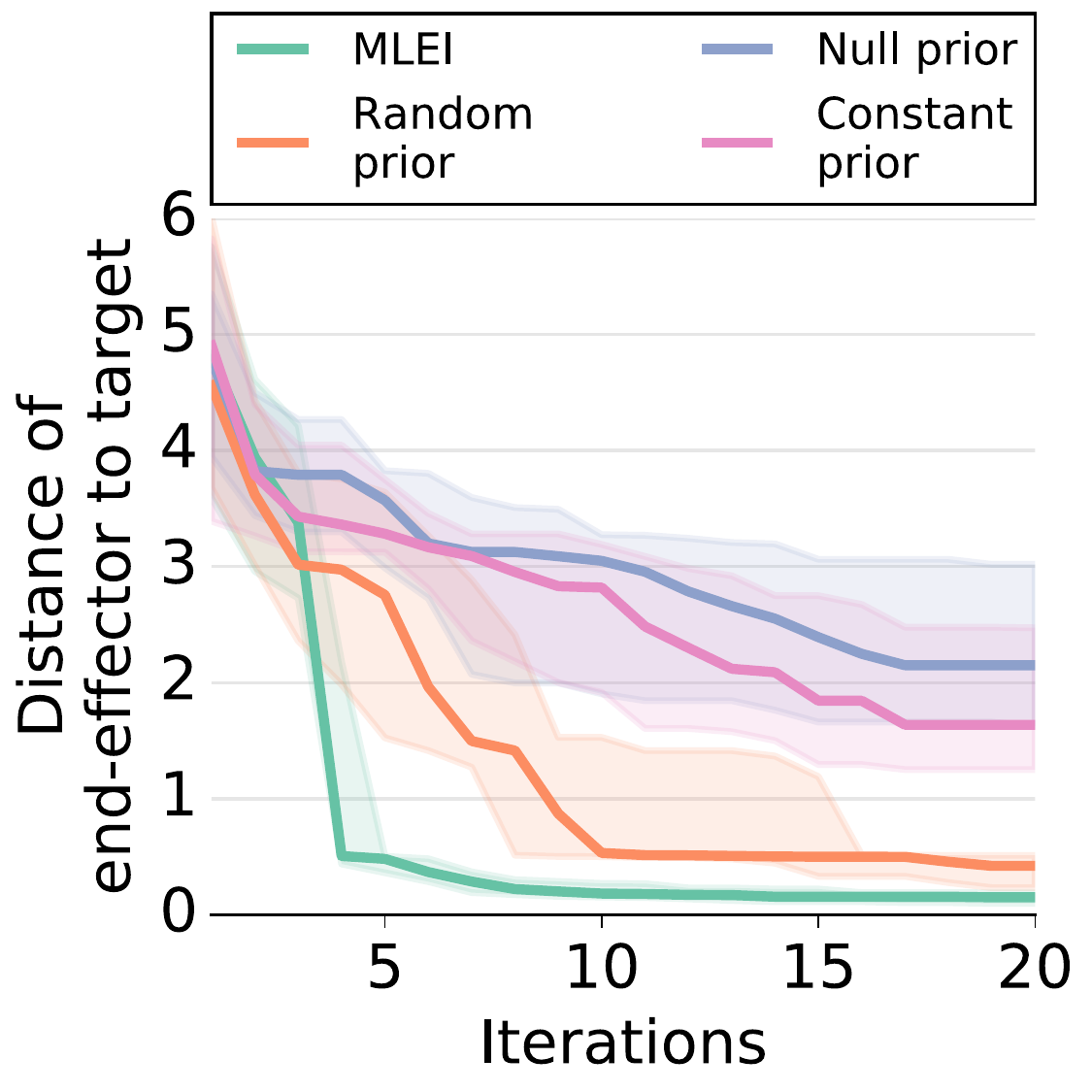}}
	\vspace{-1.5em}
	\caption{Robotic arm experiment. \textbf{(a)} A 5-DOF planar arm has to touch a given target with its end-effector. The red circles correspond to the target points of the available priors, whereas the green cross indicates the actual target. A solution found by MLEI is shown. \textbf{(b)} Comparison of the MLEI acquisition function with EI without prior (traditional BO), EI with a constant prior mean function and EI with a random selection of priors. $30$ replicates of the experiment have been carried out, each one of them consisting in $20$ iterations of BO (including $3$ initial random trials).}
	\label{arm_experiment}
	\vspace{-2em}
\end{figure}

We first evaluate the MLEI acquisition function with a kinematic simulation of a planar robotic arm that has to reach a target point with its end effector (Fig.\ref{arm_setup}). The arm has 5 Degrees of Freedom (DOFs) and each link measures $1$m. The reward function is the distance between the end effector and the target point $[3,3]$ (we use a negative distance because our implementation of BO maximizes the reward). The robot is position controlled and the joints can take positions in $[-\pi;\pi]$. The policy is parametrized by the 5 target angles.

We pre-defined 10 priors (\emph{i.e.}, $10$ mean functions $\mathcal{P}(\bm{x})$) using the function $\textrm{FWD}(\bm{x})$, which gives the position of the end effector given the angular positions $\bm{x}$ and the forward kinematics:
\begin{itemize}
\item the null prior: $\mathcal{P}_1(\bm{x}) = 0$ (\emph{i.e,} the traditional BO algorithm);
\item $\mathcal{P}_2(\bm{x}) = - \Vert \textrm{FWD}(\bm{x})  - \bm{y}_2 \Vert, \bm{y}_2 = [3.6, 3.3] $ (this corresponds to a good prior since the point $\bm{y}_2$ is close to the actual target);
\item $\mathcal{P}_3(\bm{x}) = - \Vert \textrm{FWD}(\bm{x})  - \bm{y}_3 \Vert, \bm{y}_3 = [2, 2] $ (fair prior);
\item $\mathcal{P}_4(\bm{x}) = - \Vert \textrm{FWD}(\bm{x})  - \bm{y}_4 \Vert, \bm{y}_4 = [0, 0] $ (not so good prior);
\item $\mathcal{P}_5(\bm{x}) = - \Vert \textrm{FWD}(\bm{x})  - \bm{y}_5 \Vert, \bm{y}_5 = [-3, -3] $ (bad prior);
\item $\mathcal{P}_{6,...,10}(\bm{x}) = - \Vert \textrm{FWD}(\bm{x})  - \bm{y}_{5+i} \Vert$ where $i = 1, ..., 5$ and $\bm{y}_{5+i}$ is randomly chosen in $[-3,3]$ (this is done once before all the experiments; see Fig.~\ref{arm_setup} for the target points used in the experiments).
\end{itemize}
None of these priors corresponds to the right target, but for instance, if the second prior is selected, then the robot ``knows'' how to reach the target at $[3.6, 3.3]$. This setup can be seen as a simple transfer learning example: (1) the robot knows how to reach some targets, that is, how to solve some tasks, and (2) the robot needs to learn how to reach a novel target given the knowledge of previous targets.

The optimization is initialized by three random trials of the robotic arm and then BO is used for 17 iterations to select the next move of the arm (for a total of 20 episodes on the robot). We compare four different variants of BO:
\begin{itemize}
\item EI with null prior: standard BO using EI without prior (the mean function is equal to $0$ --- this is an optimistic prior \cite{lizotte2005gaussian});
\item EI with constant prior: BO using EI with constant prior (the mean function is equal to $-7$ --- this is a pessimistic prior \cite{lizotte2005gaussian});
\item EI with a prior randomly selected among the available priors at each iteration of BO;
\item MLEI with automatic selection of priors at each iteration of BO.
\end{itemize}

We replicated each experiment 30 times to gather statistics.

The results show that MLEI finds a policy that reaches the target (distance to the end effector inferior to $15cm$) after $7-8$ episodes, whereas the EI with random selection of priors and the EI with no prior need more than $20$ (Fig.~\ref{arm_results}). Overall, MLEI clearly outperforms the three baselines.

\subsection{6-legged robot experiment (adaptation to new environments and to damage)}
We then evaluate the MLEI acquisition function in a similar context as in Cully \emph{et al.}~\cite{NatureArticle}: a 6-legged robot is either damaged in an unknown way or introduced to an unknown environment and BO is used to find an alternative walking gait that works in spite of the unforeseen situation. However, while Cully \emph{et al.} used a single prior (walking on a flat surface with an intact robot), we introduce many other priors that correspond either to potential damages (\emph{e.g.}, a missing leg) or to different terrains (\emph{e.g.}, stairs). We test the learning algorithm with priors corresponding to the actual situation, but also in situations that are not fully covered by any prior.

\subsubsection{Robot and policy}
\begin{figure}
		\centering
		\subfigure[]{\label{hexapod_simu} \includegraphics[width=0.26\textwidth]{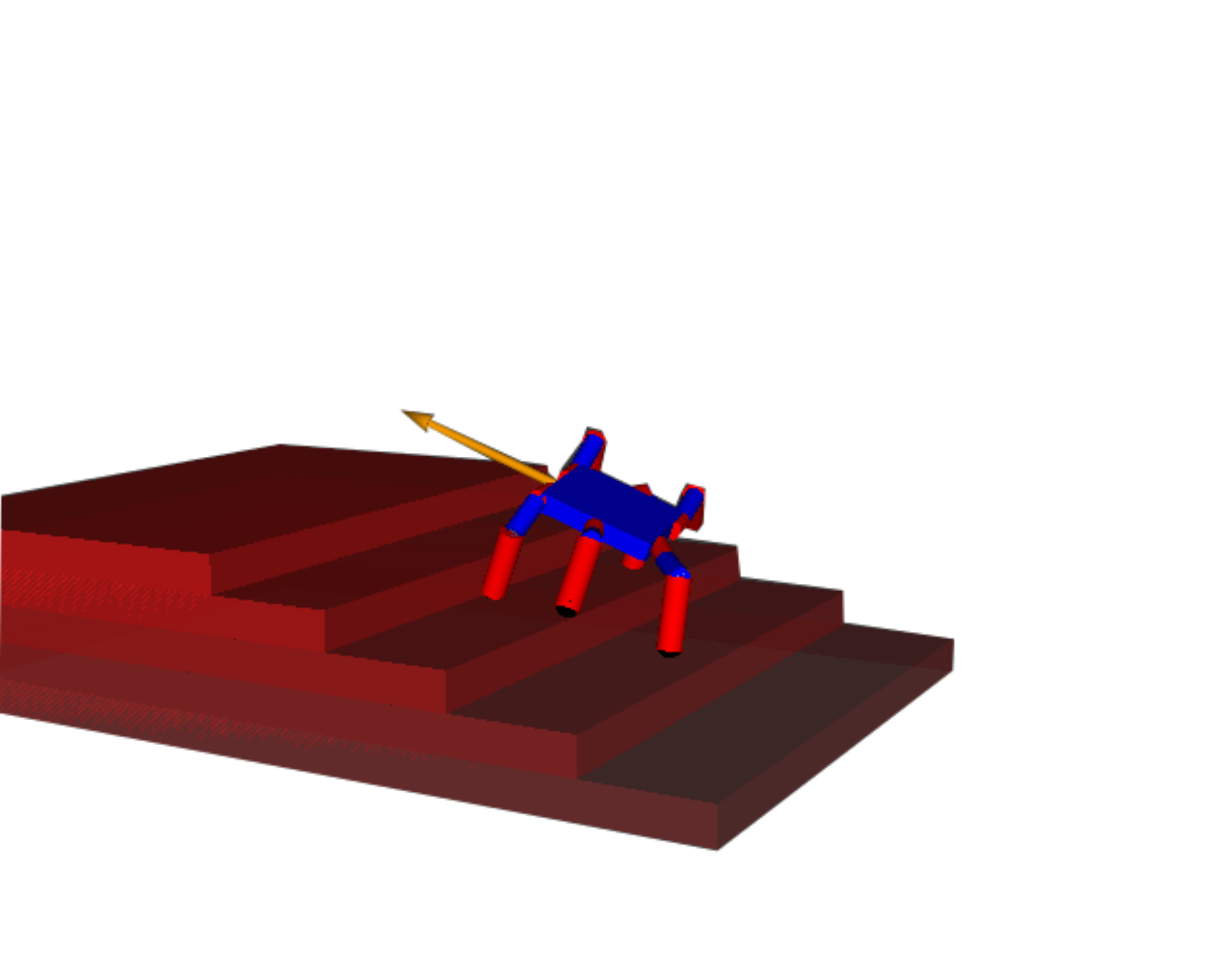}}
		\subfigure[]{\label{hexapod_real} \includegraphics[width=0.2\textwidth]{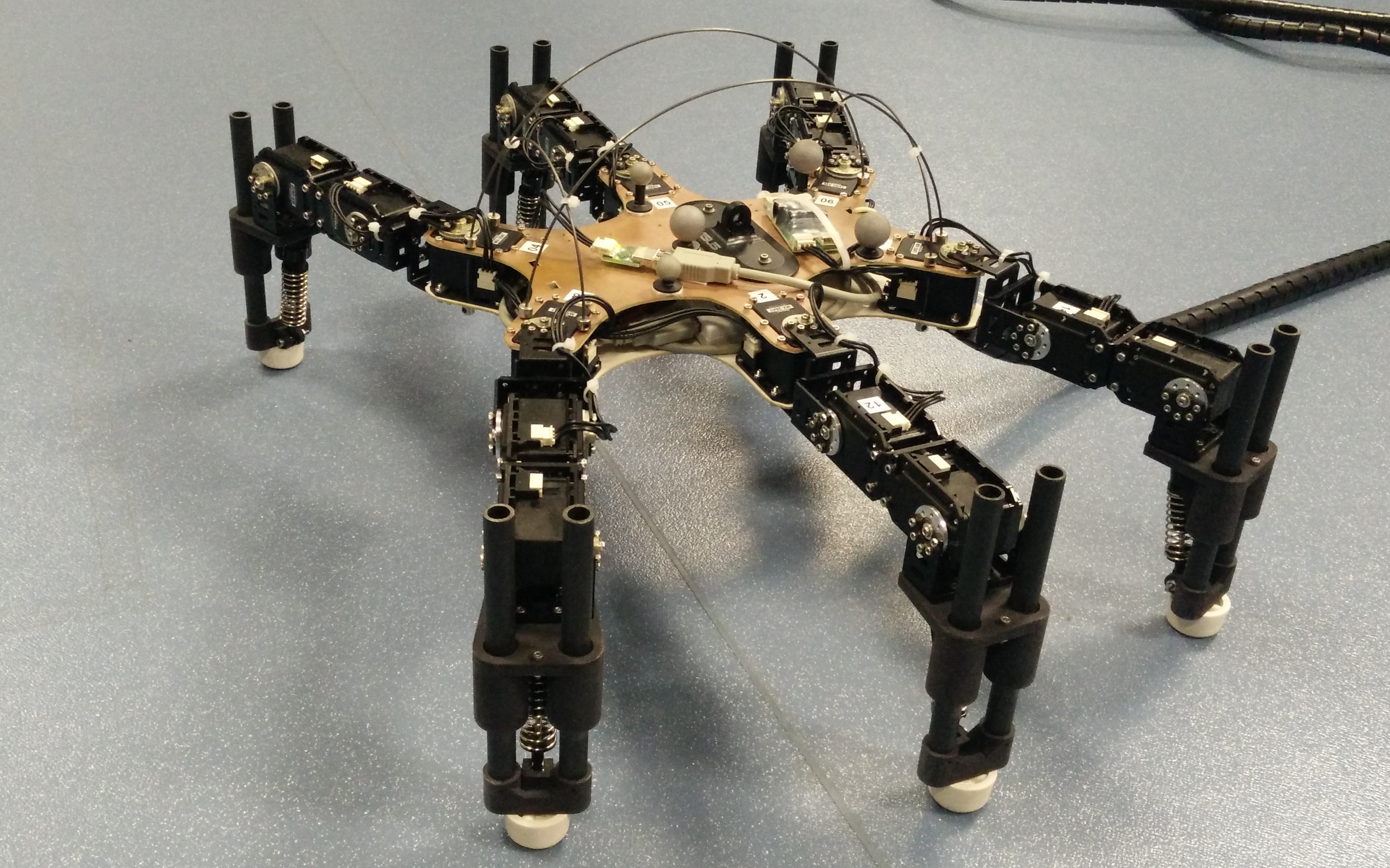}}
		\vspace{-1.5em}
    	\caption{The 6-legged robot used in the experiments: simulation of the hexapod on stairs~\subref{hexapod_simu} and real robot~\subref{hexapod_real}.}
		\label{hexapod_view}
		\vspace{-2em}
	\end{figure}

The robot has identical legs with 3 DOFs per leg (Fig.~\ref{hexapod_real}). One DOF ($\theta_1$) controls the horizontal movements of the leg (from back to front) whereas the two others ($\theta_2 \text{ and } \theta_3$) control the elevation of the leg.  Each one of these DOFs is controlled by an open-loop oscillator defined with 3 parameters \cite{NatureArticle}: an amplitude, a phase, and a duty cycle (proportion of time in which the angle is in an extreme position). The second vertical angle $\theta_3$ is constrained to take values between $-\theta_2$ and $-\theta_2 + \frac{\pi}{4}$, so that the inferior member (the "tibia") remains vertical or at most at an angle of $\frac{\pi}{4}$ with the vertical line. Thus, the whole gait of the robot can be defined with $6 \times 3 \times 3 = 54$ parameters. All simulations\footnote{https://github.com/resibots/robot\_dart} of the robot are performed with the Dynamic Animation and Robotics Toolkit (DART)\footnote{http://dartsim.github.io/} in a world with gravity were the simulated robot is similar to the intact, physical hexapod.

\subsubsection{Reward function} In all the experiments, the reward function is the distance covered by the 6-legged robot in a virtual corridor with a width of $1 m$ (the width of the robot is about $40 cm$). As soon as the robot gets out of the corridor, the evaluation is stopped; it is also stopped after $10$s if the robot stays in the corridor. Compared to more traditional reward functions, for instance the distance covered in $10$ seconds, our reward function encourages more the robot to follow a straight line, even if it means that the gait is slower. Similar results were however obtained with the average walking speed as a reward.

\subsubsection{Prior generation}
All the priors are 6-dimensional behavior-reward maps computed for a simulated 6-legged robot in different environments or with the damaged robot (\emph{e.g.}, with a missing leg).
These behavior-reward maps are created beforehand using the MAP-Elites algorithm~\cite{NatureArticle, mouret2015illuminating}, which is a recent evolutionary algorithm designed to generate thousands of different high-performing control policies\footnote{This new family of search algorithm is called ``illumination algorithms''~\cite{mouret2015illuminating} or ``quality diversity algorithms''~\cite{pugh2016quality}.}.

MAP-Elites assumes that we can define a low-dimensional \emph{behavior descriptor} for each policy, that is, a low-dimensional vector that captures the main difference between two behaviors. Given a $n$-dimensional behavior space, MAP-Elites defines a $n$-dimensional grid divided in cells, and attempts to fill each of the cells with high-quality solutions. To do so, it starts with $G$ random policy parameters, simulates the robot with these parameters, and records both the position of the robot in the behavior space and the performance. If the cell is free, then the algorithm stores the policy parameters in that cell; if it is already occupied, then the algorithm compares the reward values and keeps only the best parameter vector.
Once this initialization is done, MAP-Elites iterates a simple loop: (1) randomly select one of the occupied cells, (2) add a random variation to the parameter vector, (3) simulate the behavior, (4) insert the new parameter vector into the grid if it performs better or ends up in an empty cell (discard the new parameter vector otherwise).

MAP-Elites can be straightforwardly parallelized and can run on large clusters before deploying the robot.
So far, it has been successfully used to create behaviors for legged robots~\cite{NatureArticle,chatzilygeroudis2016reset}, wheeled robots~\cite{duarte2017evolution, pugh2016quality}, designs for airfoils~\cite{gaier2017feature}, morphologies of walking ``soft robots''~\cite{mouret2015illuminating}, and adversarial images for deep neural networks~\cite{nguyen2015deep}. MAP-Elites has also been extended to effectively handle tasks with spaces of arbitrary dimensionality~\cite{vassiliades2017using}.

We use one of the behavior descriptors proposed in Cully \emph{et al.}~\cite{NatureArticle}: the body orientation, which captures how often the body of the robot is tilted in each direction\footnote{Similar results were obtained with other behavioral descriptors.}. More formally, simulating each policy leads to a 6-dimensional vector that contains the proportion of time that the body of the robot has a positive and negative pitch, yaw and roll:

\begin{equation}
	\bm{BOF} = \begin{bmatrix}
		\frac{1}{K} \sum_{k=1}^K \mathds{1}_{\Theta(k) > 0.005\pi} \\
		\frac{1}{K} \sum_{k=1}^K \mathds{1}_{\Theta(k) < -0.005\pi} \\
		\frac{1}{K} \sum_{k=1}^K \mathds{1}_{\Psi(k) > 0.005\pi} \\
		\frac{1}{K} \sum_{k=1}^K \mathds{1}_{\Psi(k) < -0.005\pi} \\
		\frac{1}{K} \sum_{k=1}^K \mathds{1}_{\Phi(k) > 0.005\pi} \\
		\frac{1}{K} \sum_{k=1}^K \mathds{1}_{\Phi(k) < -0.005\pi}
	\end{bmatrix}
\end{equation}

where the duration of the gait of the robot is divided in $K$ intervals of 15 ms, $\Theta$, $\Psi$ and $\Phi$ are the pitch, roll and yaw of the torso of the robot, respectively, $\mathds{1}$ is the indicator function which returns 1 if and only if its argument is true, and angles between $\pm 0.005\pi$ are ignored.

This quantity is rounded so that it can only take values in $\{0, 0.2, 0.4, 0.6, 0.8\}$ and so the set of all the body orientation factors is finite and contains $5^6=15625$ elements that can be organized in a map.

For the purpose of the experiments, 15 behavior-performance maps have been created for each of the possible environments (priors). Each one of these maps was created with a run of the MAP-Elites algorithm for 24 hours on a 16-core Xeon computer. We used the Sferes C++ library~\cite{Mouret2010}.

 The kernel chosen for the GP is the Squared Exponential Kernel: $k_{SE}(\bm{x}, \bm{x'}) = \sigma^2 \exp \Big(-\frac{1}{2}(\bm{x}-\bm{x'})^T\bm{M}(\bm{x}-\bm{x'})\Big)$  where $\bm{M} = \mathrm{diag}(l_1^{-2}, \dots, l_D^{-2})$ is the characteristic length scales (here $D=6$)~\cite{Brochu}~\cite{Rasmussen}. Initially, $\sigma=1$ and $\forall i \in \{1,... ,D\}, l_i=1$ and $\sigma_n=0.00001$. The hyperparameters of the kernel are optimized through Resilient backPROPagation (RPROP)~\cite{Blum2013OptimizationOG}, with 300 iterations.

\subsubsection{\textbf{Experiment 1} --- Adaptation to stairs in simulation}
In our first set of experiments, the intact robot needs to adapt to unknown environments. We generated $15$ behavior-performance maps (\emph{i.e.}, 15 priors for the GP) for each of the four following environments:

\begin{figure*}
	\centering
	\includegraphics[width=0.85\textwidth]{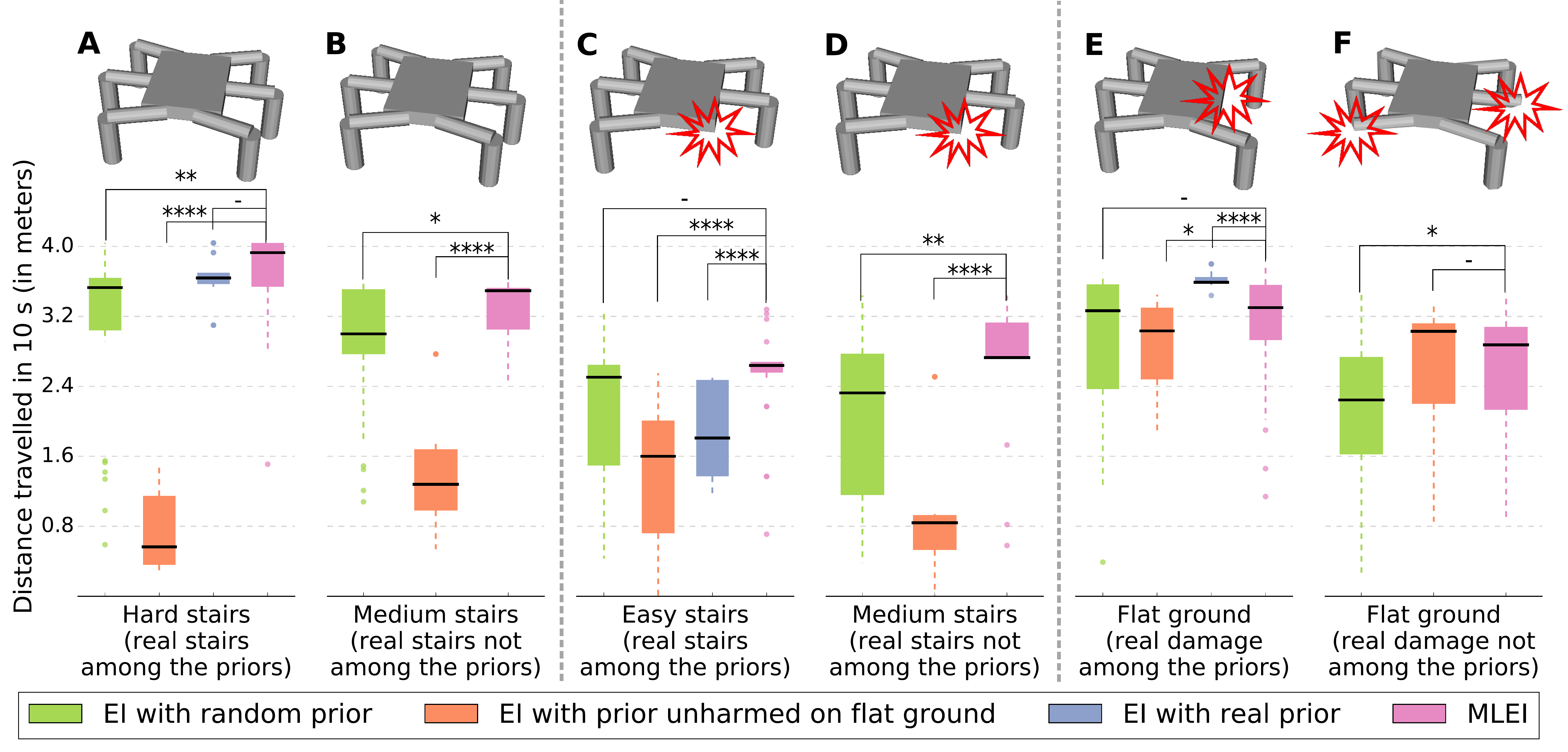}
	\vspace{-1em}
	\caption{Comparison in simulation of MLEI with other acquisition functions and choices of prior (EI with a randomly selected prior, EI with a prior generated on an unharmed robot on flat ground and EI with the prior corresponding to the real stairs or damage) on the 6-legged robot learning to climb stairs and/or to recover from damages after 5 iterations of BO and with 30 replicates of each experiment. (A) and (B): the robot is on unknown stairs with no damage and the real stairs can be among the priors (A) or not (B). (C) and (D): the robot is on unknown stairs with unknown damages and the priors are only on stairs not on damages (the actual stairs can be among the priors (C) or not (D)). (E) and (F): the robot is on flat ground with unknown damages and the real damage can be among the priors (E) or not (F). The number of stars indicates that the p-value, obtained using the Mann-Whitney-Wilcoxon test, is below 0.0001, 0.001, 0.01 and 0.05 respectively.}
	\label{exp_simulation}
	\vspace{-2em}
\end{figure*}

\begin{itemize}
\item flat ground;
\item easy stairs (steps with height: $4cm$, width: $1.2 m$, depth: $50 cm$);
\item medium stairs (steps with height: $5 cm$, width: $1.2 m$, depth: $20 cm$);
\item hard stairs (steps with height: $7.5 cm$, width: $1.2 m$, depth: $25 cm$).
\end{itemize}

\noindent We compare the following acquisition functions for BO:
\begin{itemize}
\item EI with a single prior coming from a simulated robot on flat ground -- this corresponds to the original IT\&E experiments~\cite{NatureArticle};
\item EI with a single prior, randomly chosen among the available priors at each iteration.
\item EI with a single prior coming from a simulated robot on the actual stairs (when available) -- this corresponds to the ideal case, in which we know the right prior;
\item MLEI with a prior selected at each iteration among the available priors (flat ground, easy, medium and hard stairs).
\end{itemize}

For the MLEI and EI with random priors experiments, we randomly choose $5$ priors (\emph{i.e.}, 5 maps) for each possible environment, leading to a unique set of priors for each MLEI experiment and for each experiment with randomly chosen priors. Please note that several priors correspond to the same situation, which is interesting because some maps might be of higher-quality than others, even if they have been generated with the same environment.

\noindent We test two situations:
\begin{enumerate}
	\item adaptation to hard stairs when the hard stairs are part of the priors given to MLEI (and to random selection) --- $5 \times 4= 20$ priors to select from;
	\item adaptation to medium stairs, with the medium stairs removed from the priors given to MLEI (and to random selection) --- $5 \times 3 = 15$ priors to select from.
\end{enumerate}
In these two situations, the robot is the same in the prior and in the adaptation experiment (there is no ``reality gap'').

The results (Fig.~\ref{exp_simulation}A-B) show that MLEI allows the robot to learn high-performing gaits for the stairs, even when the stairs used for the learning experiments are not present in the set of priors (Fig.~\ref{exp_simulation}B): when the right prior is accessible, MLEI finds it; when it is not accessible, it can still leverage other priors and use BO to find a good behavior while using other priors. In the two tested cases, MLEI clearly outperforms the random selection of priors and the method using the flat ground prior, which means that MLEI selects priors correctly and that these priors help the learning process. Surprisingly, MLEI also outperforms the EI with a ``perfect'' prior (Fig.~\ref{exp_simulation}A): this is because MLEI has access to $5$ priors for the hard stairs (in addition to the $15$ other priors) and therefore can select the best of them, whereas each EI experiment has access to a single prior for the considered stairs (and the best controller for each map is different). The relatively good performance of the random selection of priors is likely to stem from the fact that this algorithm has access to a much higher diversity of behaviors than EI with flat ground as a prior (that is, to the original IT\&E), which makes it more likely to find a behavior that works in the tested situation.

 \subsubsection{\textbf{Experiment 2} --- Adaptation to stairs and damages in simulation}
 In this second experiment, we evaluate if the robot can adapt to unforeseen damage conditions, with and without stairs, with and without priors about the damage conditions. For each of the 6 legged removed, we generated $15$ priors with MAP-Elites (with a robot on flat ground), leading to $(6+1) \times 15 = 105$ priors (6 damage conditions + intact robot). Like in the previous experiments, the set of available priors is made of $5$ random maps (out of the $15$ generated priors) for each situation.

 We compare the same methods as before in four situations that cover different combinations of environmental and body-related priors:
 \begin{itemize}
	\item[1.a] adaptation to damage with priors about stairs (no prior about damage), and when the actual stairs are among the priors --- 20 priors to select from;
	\item[1.b] adaptation to damage with priors about stairs (no prior about damage), and when the actual stairs are not among the priors --- 15 priors to select from;
	\item[2.a] adaptation to damage with priors about the damage conditions, on flat ground, when the actual damage (left middle leg removed) is among the priors --- $7 \times 5=35$ priors to select from;
	\item[2.b] adaptation to damage with priors about the damage conditions, on flat ground, when the actual damage (front right leg and middle left leg shortened) is not among the priors --- $7 \times 5 = 35$ priors to select from.
 \end{itemize}

The results (Fig.~\ref{exp_simulation}C-F and supplementary video\footnote{\label{video}Also at: {\scriptsize\url{https://youtu.be/xo8mUIZTvNE}}}) show that MLEI can find compensatory gaits on stairs while using priors computed with the intact robot. When the real stairs are among the priors (Fig.~\ref{exp_simulation}C), MLEI outperforms the EI with the right stairs because (1) since the robot is damaged, the most helpful prior is not always the prior that corresponds to the correct stairs (\emph{e.g.}, the prior that corresponds to the hard stairs might be more conservative and be more helpful when the robot is damaged); (2) like before, MLEI has access to more priors, which makes it more likely to have a policy in one of the map that can compensate for the damage.

When the actual stairs are not in the priors, MLEI still outperforms the two other approaches (Fig.~\ref{exp_simulation}D).  MLEI can also take advantage of priors about the damage condition whether the damage is included in the priors or not (Fig.~\ref{exp_simulation}E-F): when the actual damage conditions is among the priors, MLEI leads to higher-performing solutions than EI with the intact robot as a prior; when the damage condition is not among the prior, MLEI performs the same as EI with the intact robot as a prior. These results are consistent with \cite{NatureArticle}, which shows that an intact robot can be an effective prior to adapt to damage.

 \subsubsection{\textbf{Experiment 3} --- Adaptation to damage with a physical robot}
In this experiment, we use $(6+1) \times 15 = 105$ priors for damage conditions to allow a physical 6-legged robot to adapt. As the simulation is not perfect, the learning algorithm has to compensate for both the ``reality gap'' and the damage. The robot is tracked with an external motion capture system (Optitrack) and we use 10 episodes of 10 seconds. Like before, we consider two situations: when the damage is among the priors, and when it is not. We replicate each experiment $5$ times.

\begin{figure}
    	\centering
		\subfigure[]{\label{missing_rear}
		\includegraphics[width=0.23\textwidth]{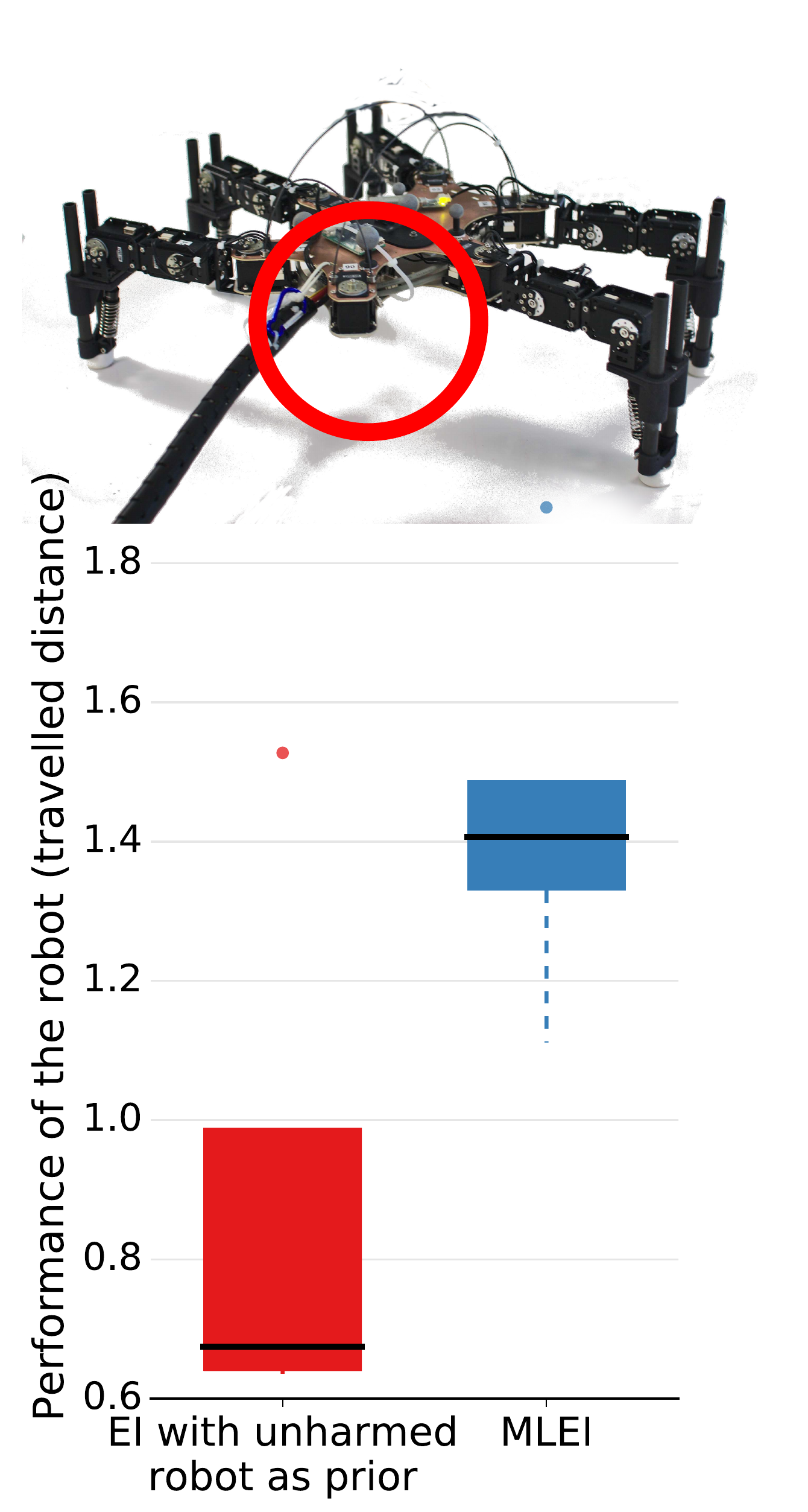}}
		\subfigure[]{\label{short_rear}
		\includegraphics[width=0.23\textwidth]{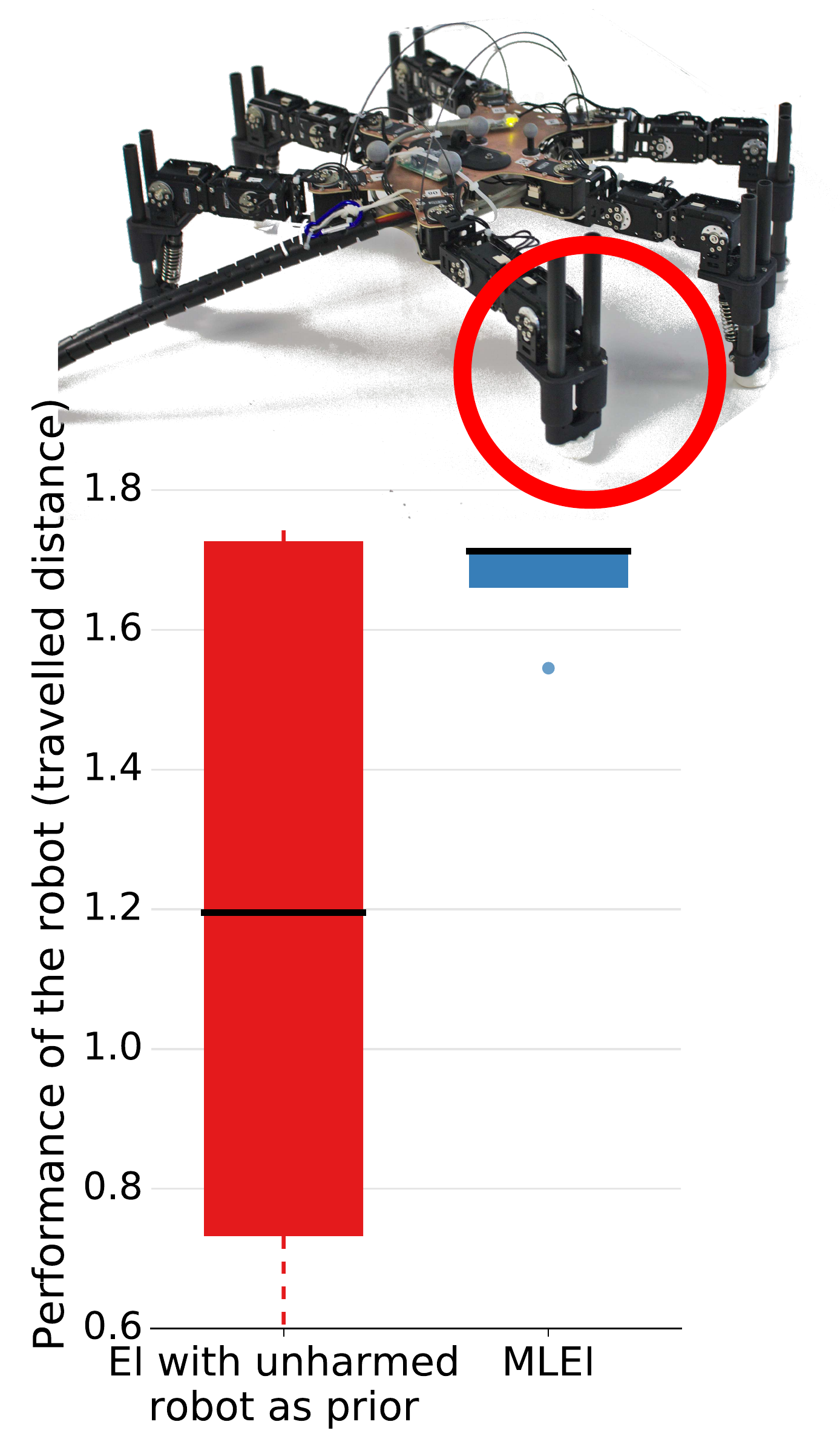}}
		\vspace{-1.5em}
    	\caption{Comparison of MLEI with the standard EI with a single prior coming from a simulated undamaged robot. This real experiment was carried out on the physical damaged 6-legged robot walking on flat ground after 10 iterations of BO and with 5 replicates of the experiment. Damage used: \subref{missing_rear} missing rear leg (damage present among the available priors), \subref{short_rear} shortened rear leg (damage not present among the available priors).}.
		\label{prior_damages_real}
		\vspace{-2em}
\end{figure}

Like in simulation, MLEI takes advantage of the priors to find higher-performing gaits than when a single prior is used (Fig.~\ref{prior_damages_real} and supplementary video\footnoteref{video}). When one of the priors correspond to the actual damage condition (Fig.~\ref{prior_damages_real}(a)), MLEI clearly outperforms EI with a single prior and finds high-performing gaits in less than 10 episodes; MLEI also finds better gaits than EI when the actual damage condition is not among the priors (Fig.~\ref{prior_damages_real}(b)), which is likely to come from the fact that MLEI can ``take inspiration'' from other priors to compensate for the damage (like in the previous task, this corresponds to a form of transfer learning).

\section{Conclusion and Discussion}
Well-chosen priors can guide BO to find a high-performing solution \cite{BOgait,NatureArticle} while not constraining the search to a few pre-designed solutions. However, learning algorithms are most useful when the robot or the environment are partially known, therefore it is often challenging to design a single prior that would help BO in all the possible situations. The Most Likely Expected Improvement (MLEI) allows us to relax this assumption by making BO capable of selecting the most useful prior and ignore all the others. It therefore makes it possible for BO to benefit from the faster convergence speed given by the priors, while not assuming much about the robot or the environment.

In this paper, we demonstrated that our new acquisition function leads to a powerful adaptation algorithm in two systems, a planar manipulator and a 6-legged robot. In the latter case, the robot was capable of discovering compensatory behaviors in a dozen of trials when damaged --- even with priors that correspond to the intact robot ---  and when it faced unknown stairs -- even without any prior for the actual stairs. Overall, MLEI substantially increases the potential uses of priors in BO because we can often ``guess'' what could be useful to the robot, but we cannot be sure in advance if a given prior will be useful in the future.

Even the best classification system based on perception (which context is recognized by the robot?)~\cite{plagemann} is prone to errors in real situations (\emph{e.g.}, steps hidden by high grass). By contrast, the automatic selection of priors that we introduced here is based on the direct observation of the rewards: the robot discovers what works and what does not, it does not attempt to know \emph{why} some behaviors work and some do not. This approach fits well the theory of ``embodied cognition'' \cite{brooks1991intelligence,pfeifer2006body} which suggests that robots do not need an explicit representation of the world to act. A classic ``sense-plan-act'' architecture would assume the existence of an accurate model of the world to act; at the other end of the spectrum, most learning algorithms aim at assuming as little knowledge as possible about the robot or the environment. BO with automatic selection of prior can be an effective middle ground in which prior knowledge or a perception system can guide a direct policy search that can, if needed, ignore all previous knowledge and still find an effective way to act.

\vspace{-0.4em}
\appendix
\vspace{-0.5em}
Code of the experiments: {\scriptsize\url{https://github.com/resibots/pautrat_2018_mlei}}
\vspace{-1em}
\bibliographystyle{IEEEtran}
\bibliography{IEEEabrv,references}

\end{document}